\documentclass[letterpaper, 10 pt, conference]{ieeeconf}  

\IEEEoverridecommandlockouts
\usepackage[utf8]{inputenc}
\usepackage[T1]{fontenc}
\usepackage{cite}
\usepackage{hyperref}
\usepackage{url}
\usepackage{booktabs}
\usepackage{amsfonts}
\usepackage{nicefrac}
\usepackage{microtype}
\usepackage{xcolor}
\usepackage{textcomp}
\usepackage{graphicx}
\usepackage{amsmath}

\newcommand{\ours}{\textbf{Interfaze}}
\newcommand{\oursbeta}{\textbf{Interfaze-Beta}}

\title{\LARGE \bf
Interfaze: The Future of AI is built on Task-Specific Small Models
}

\author{
\authorblockN{Harsha Vardhan Khurdula}
\authorblockA{
JigsawStack, Inc. \\
San Francisco, CA, USA \\
harsha@jigsawstack.com
}
\and
\authorblockN{Vineet Agarwal}
\authorblockA{
JigsawStack, Inc. \\
Durgapur, WB, India \\
vineet@jigsawstack.com
}
\and
\authorblockN{Yoeven D Khemlani}
\authorblockA{
JigsawStack, Inc. \\
San Francisco, CA, USA \\
yoeven@jigsawstack.com
}
}

\begin{document}

\maketitle
\thispagestyle{empty}
\pagestyle{empty}

\begin{abstract}
We present \ours{}, a native hybrid model that fuses task-specific deep neural networks (CNNs and DNNs) directly into a transformer decoder through a shared embedding space. Specialized perceptual encoders handle optical character recognition (OCR) over complex multilingual PDFs, open-vocabulary object and graphical user interface (GUI) detection, and multilingual speech recognition with diarization. Each is exposed through a task-specific adapter and can be activated on its own, so a query touches only the parameters it needs. A built-in action foundation supplies a grounded external state: a proxied headless browser and scraper, a code sandbox, a multi-domain web index, and a scalable vector store. The decoder filters and merges these signals, reasons over them when a task requires it, and emits deterministic outputs built on confidence. The raw specialist metadata (bounding boxes, confidence scores, timestamps) is preserved and returned alongside the answer as \emph{precontext}.

On this architecture, \oursbeta{} leads a suite of deterministic developer-task benchmarks. It reaches \textbf{70.7\%} on OCRBench~v2, \textbf{85.7\%} on olmOCR, \textbf{82.1\%} on RefCOCO, a \textbf{2.4\%} word error rate on VoxPopuli, \textbf{52.9\%} on Spider-2.0-Lite, \textbf{92.4\%} on GPQA-Diamond, \textbf{90.9\%} on MMMLU, \textbf{71.1\%} on MMMU-Pro, and \textbf{80.5\%} value accuracy on the Structured Output Benchmark (SOB), ahead of comparably priced generalist models (Gemini-3-Flash, Gemini-3.5-Flash, Claude-Sonnet-4.6, GPT-5.4-Mini, and Grok-4.3) on every task. Because fused specialist encoders resolve perception in a single pass instead of through repeated tool calls into a large model, \ours{} reaches high accuracy with verifiable metadata on deterministic tasks while running at flash-tier cost.
\end{abstract}

\section{Introduction}

Many production AI workloads are not open-ended generation but deterministic perception and extraction: reading complex documents, detecting objects and interface elements, transcribing and diarizing speech, translating, scraping the live web, and returning structured records. These tasks reward precision and verifiable metadata (bounding boxes, calibrated confidence, timestamps) rather than fluent prose, and they run at high volume, where cost and latency matter.

Two families of models each solve half of the problem. Classical CNN and DNN specialists produce pixel-tight boxes, calibrated per-output confidence, and deterministic results at millisecond cost, but they break on formats they were not trained on, need expensive retraining, and have little semantic flexibility \cite{sanh2019distilbert, sun2020mobilebert}. Monolithic transformers and vision-language models bring flexibility, reasoning, and schema following, but their boxes are imprecise, their confidence is a token probability rather than a visual one, they hallucinate on degraded inputs, and they are costly to run \cite{belcak2025slm_agents}.

A common remedy is to let a large model route between models or call specialist tools \cite{chen2023frugalgpt, ding2024hybrid, jitkrittum2025uniroute, schick2023toolformer, shen2023hugginggpt, yao2022react, chameleon2023}. This recovers some accuracy, but the large model stays the final author and can paraphrase or drop details while formatting; it cannot tell when a specialist misread a character or skipped a line; bounding boxes are passed through a text channel and corrupted; every pass adds latency; and the routing decision itself is unpredictable. Tool calling is useful for orchestration, but it is the wrong pattern when a specialist's output must reach the response unchanged and the two models need to check each other.

We take a different approach. Instead of selecting among separate models, we fuse task-specific CNN and DNN encoders directly into a transformer decoder through a shared embedding space. The decoder reads specialist representations and their metadata together with its own semantic tokens, and can reconcile the two in a single pass, for example by correcting an OCR misread against the surrounding context or by grounding a generated field in a detected box. This builds on a line of work that bridges modality-specific encoders into language models \cite{flamingo, blip2, llava, metachameleon}, specializes a backbone with lightweight adapters \cite{lora}, and activates only part of a model for a given input \cite{moe_shazeer}.

\ours{} has three parts:
\begin{enumerate}
  \item a heterogeneous stack of CNN and DNN perceptual encoders, fused with a transformer backbone through task adapters in a shared embedding space, covering OCR and layout, object and GUI detection, multilingual ASR with diarization, translation, and lightweight classification and statistical prediction, with support for partial activation;
  \item a built-in action and infrastructure foundation, trained into the model rather than attached as external tools, comprising a proxied headless browser with a self-healing scraper, a code sandbox, a multi-domain web index (general search, financial, and research sources), and a scalable vector store with caching; and
  \item a decoder that filters and merges these signals, reasons over them when needed, and produces developer-defined structured output through a validation layer, while preserving the specialist metadata as \texttt{precontext} and offering a partial-activation path that returns a single task's fixed-schema output without running the full decoder.
\end{enumerate}

We use \ours{} for the architecture and \oursbeta{} for the model that instantiates and is evaluated in this paper, and refer to them this way throughout. Our concrete instantiation, \oursbeta{}, fuses specialized encoders for the major deterministic tasks behind a single chat completion compatible endpoint. We describe each component and evaluate the system on OCR, document parsing, visual grounding, speech, text-to-SQL, multimodal and multilingual understanding, and structured output.

We contribute the following:
\begin{itemize}
  \item a native hybrid architecture that fuses task-specific CNN and DNN encoders into a transformer decoder through a shared embedding space and task adapters, so that perception is part of the model rather than an external tool;
  \item a two-tier design that separates fused perceptual encoders, which are resolved inside the decoder from a built-in action foundation (browser, scraper, sandbox, web index, and vector store) whose results are merged into the decoder and returned to developers as \texttt{precontext};
  \item a partial-activation mechanism that runs only the parameters a task needs and can bypass the decoder to return a single task's fixed-schema output;
  \item \oursbeta{}, an instantiation that leads deterministic-task benchmarks against comparably priced generalists while returning first-class metadata; and
  \item an analysis of where native fusion falls short, chiefly encoder coverage and retraining cost, with directions for future work.
\end{itemize}

\section{Literature Survey}

\subsection{Context is key}

\begin{table}[h]
  \centering
  \caption{Average long-context benchmark performance for several LLMs (adapted from Hron~\cite{hron2025legalcontext}).}
  \label{tab:long_context_scores}
  \begin{tabular}{lcc}
    \hline
    Model & 1M ctx.\ score & 128K ctx.\ score \\
    \hline
    gpt-4o-2024-11-20      &  50 & 60 \\
    gpt-4.1-alpha          &  55 & 75 \\
    o3-mini-2025-01-31     &  50 & 55 \\
    o1-2024-12-17          &  56 & 62 \\
    claude-3-7-sonnet      &  50 & 58 \\
    claude-3-7-sonnet-thinking & 54 & 63 \\
    gemini-1.5-pro-002     &  56 & 67 \\
    \hline
  \end{tabular}
\end{table}

Recent work suggests that \emph{how} context is selected and organized matters more than simply enlarging a model or its window. A Databricks study on financial and corporate QA varies both retrieved document count and prompt length for long-context models such as GPT-4 Turbo and Claude 2, and observes gains only while the retrieved text remains dense and relevant; once loosely related passages dominate, quality degrades even when the full context fits \cite{leng2024longcontext}. In legal QA, Hron reports a similar pattern: feeding entire filings sometimes beats naive RAG over small snippets, but performance still drops on very long documents with dispersed evidence \cite{hron2025legalcontext}. Xu et al.\ show that moderate context windows combined with retrieval can match or outperform larger long-context baselines as shown in Table~\ref{tab:long_context_scores} across knowledge and reasoning benchmarks \cite{xu2023retrieval}. Together, these results argue that context must be filtered, structured, and budgeted. We extend this view: rather than assembling context for a separate LLM, \ours{} fuses compact perceptual representations into the decoder, so structured and budgeted signals reach the model without a separate retrieval and prompting stage.

\subsection{Where routing falls short}

Hybrid routing work asks \emph{which} LLM to call, and when a smaller model is ``good enough.'' FrugalGPT proposes cost-aware cascades over proprietary models \cite{chen2023frugalgpt}; Hybrid LLM adds a router that predicts query difficulty and routes between a local small model and a larger cloud model \cite{ding2024hybrid}; Universal Model Routing generalizes this idea to changing expert pools \cite{jitkrittum2025uniroute}; and routed architectures show that many experts can remain idle unless explicitly activated \cite{clark2022unified}. In almost all cases the experts are text-only LLMs of different sizes and prices. Vision, speech, OCR, and retrieval are treated as fixed utilities, if they are modeled at all. This leaves open how to route over a heterogeneous DNN stack, or how much quality and cost are driven by perception rather than by the final LLM. We avoid routing entirely. Rather than choosing among separate models or tool chains, \ours{} fuses the specialist encoders into the decoder, so perception and generation take place in a single model, and the decoder both reads the specialist signals and writes the final structured output.

\subsection{Tools and model composition}

A second line of work studies LLMs that call external tools. Toolformer demonstrates that a large model can label its own training data with API calls and learn when to invoke calculators, search, or translation systems \cite{schick2023toolformer}. HuggingGPT treats a large model as a planner over a registry of specialist models hosted on Hugging Face \cite{shen2023hugginggpt}. Prompting patterns such as ReAct interleave reasoning steps with explicit tool calls \cite{yao2022react}, and frameworks like Chameleon treat models and tools as composable modules arranged in small pipelines or trees \cite{chameleon2023}. Surveys summarize emerging design patterns and failure modes in such systems \cite{shen2024llmtools}. In most of this work, tools are defined only by high-level function signatures; the underlying DNNs for OCR, chart and diagram parsing, ASR, and retrieval are abstracted away. This makes it difficult to see which parts of the perception stack matter for benchmarks such as AI2D, ChartQA, MMMU, or Common Voice, where reliably reading diagrams, charts, and speech is the core challenge \cite{kembhavi2016ai2d,masry2022chartqa,yue2024mmmu,ardila2020commonvoice}. \ours{} departs from the tool-call interface for perception. OCR, detection, and ASR specialists are fused into the decoder through a shared embedding space rather than invoked as opaque functions, so their outputs and metadata are available to the model directly. Action capabilities (a browser, a scraper, a sandbox, a web index, and a SQL store) are instead trained into the model and merged into its generation, rather than orchestrated by an external planner.

\begin{figure*}[h]
  \centering
  \includegraphics[width=\textwidth]{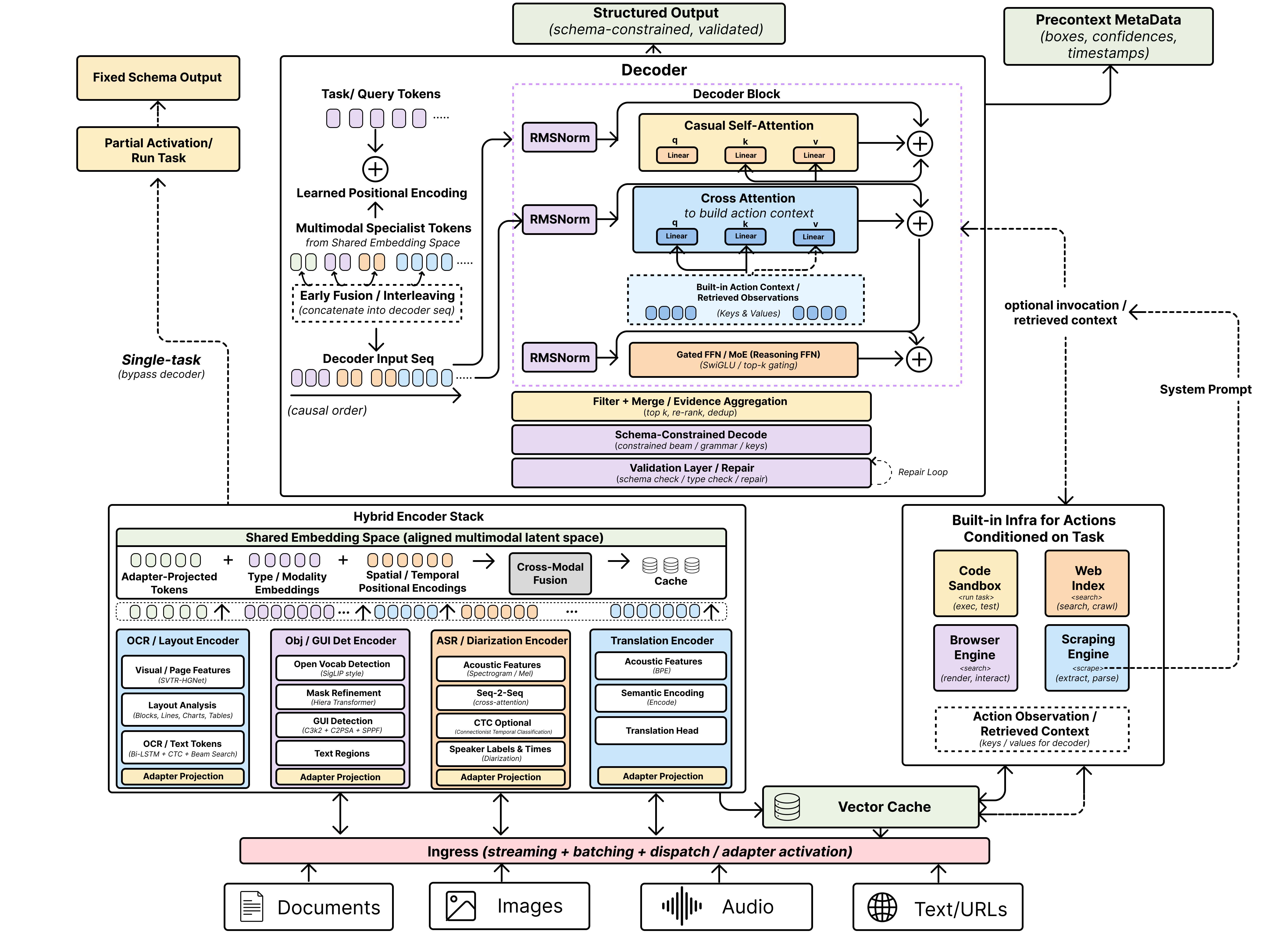}
  \caption{The \ours{} native hybrid architecture. Task-specific CNN and DNN
  encoders project into a shared embedding space and are interleaved (early
  fusion) into the decoder's causal self-attention sequence, while the built-in
  action foundation (sandbox, web index, browser, scraper) supplies retrieved
  observations to the decoder through optional cross-attention. Signals are
  filtered and merged, decoded under schema constraints, and validated into
  developer-defined structured output, with specialist metadata returned as
  \texttt{precontext}. A partial-activation path bypasses the decoder to return a
  single task's fixed-schema output.}
  \label{fig:arch}
\end{figure*}

\subsection{Small language models and specialization}

A complementary thread argues that small language models (SLMs) can be effective specialists. Schick and Sch{\"u}tze show that small models can act as strong few-shot learners on narrow tasks with the right prompting and calibration \cite{schick2020small}, \cite{easyocr2020}. Subsequent work uses compact models as plug-in components for retrieval and ranking, improving larger systems on targeted subtasks even when the small models are not competitive as general chat models \cite{acl2024findings18}. Flipping Knowledge Distillation uses small domain experts to transfer their strengths into larger generalists \cite{li2025flippingkd}, and recent position papers argue that for many agentic or tool-heavy workflows, small models are attractive due to lower latency and energy use \cite{belcak2025slm_agents}. Classic compression work such as DistilBERT and MobileBERT makes the same argument for mobile and edge deployment \cite{sanh2019distilbert,sun2020mobilebert}. Less explored is how to organize systems in which SLMs handle most of the perception and context-building work across modalities, rather than just classification or reranking. \ours{} addresses this gap. In \oursbeta{}, compact CNNs and DNNs handle OCR and document layout, object and GUI detection, speech recognition and diarization, translation, and lightweight classification, but instead of feeding a separate LLM they are fused into the decoder, so a single model perceives and generates together.

\subsection{Multimodal fusion, adapters, and conditional computation}

A growing body of work fuses modality-specific encoders into language models. Flamingo bridges frozen vision encoders to a language model with gated cross-attention \cite{flamingo}; BLIP-2 maps a frozen image encoder into a frozen language model through a lightweight querying transformer \cite{blip2}; LLaVA aligns visual features into the token space with a simple projection \cite{llava}; and Chameleon trains a single transformer over mixed-modal tokens in one early-fusion space \cite{metachameleon}. ImageBind shows that several modalities can share one embedding space \cite{imagebind}, and SALMONN connects speech and audio encoders to a language model \cite{salmonn}; production systems such as Qwen2-VL follow the same encoder-into-decoder recipe at scale \cite{qwen2vl}. A separate line of work specializes a backbone cheaply: bottleneck adapters insert small per-task modules \cite{houlsby2019adapter}, and LoRA adds low-rank task updates \cite{lora}. Conditional computation activates only part of a model for each input, from the sparsely gated mixture-of-experts layer \cite{moe_shazeer} to Switch Transformers \cite{switch_transformer} and Mixtral \cite{mixtral}. For documents, specialist visual encoders such as Nougat read academic PDFs end to end \cite{nougat}, and self-supervised speech encoders such as wav2vec~2.0 underpin modern ASR \cite{wav2vec2}. \ours{} draws these threads together: task-specific CNN and DNN encoders are fused into a transformer decoder through a shared embedding space, specialized by task adapters, activated in part for each query, and complemented by a built-in action foundation, so that one model both perceives and generates.

\section{Interfaze-Beta Architecture}
\label{sec:architecture}

\oursbeta{} is a single model organized into four stages, shown in Figure~\ref{fig:arch}: an ingress stage; a hybrid perceptual-encoder stack exposed through task adapters; a built-in action and infrastructure foundation; and a decoder that filters, merges, reasons when needed, and emits validated structured output. A partial-activation path lets a single task bypass the decoder.

\begin{figure*}[h]
  \centering
  \includegraphics[width=\textwidth]{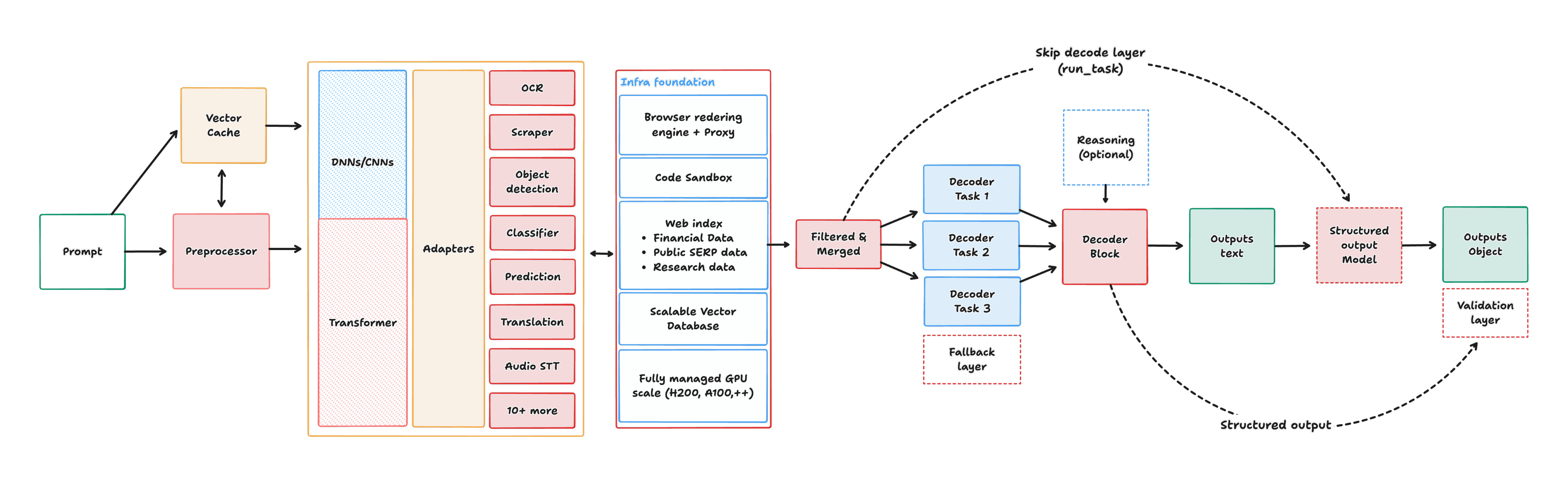}
  \caption{High level depiction of Interfaze architecture showcasing the flow for user inputs.}
  \label{fig:system_overview}
\end{figure*}

\subsection{System overview}

A request $x$ may contain text, images, audio, documents, or references such as URLs. The ingress stage, shown at the left of Figure~\ref{fig:arch}, normalizes the input, detects its modalities and intent, and runs fast safety checks \cite{llamaguard4}, while a vector cache short-circuits work that has been computed before. Ingress then activates the task adapters that the request needs over a shared backbone of CNNs, DNNs, and a transformer.

A router would dispatch the request to separate models. \ours{} instead has the perceptual encoders write into the same embedding space the transformer operates on, so their outputs and metadata are available to the decoder directly rather than as the result of a tool call. The specialist encoders carry most of the compute for deterministic tasks, and the decoder reads their compact representations and produces the answer in a single pass.

\subsection{Hybrid perceptual-encoder stack and task adapters}
\label{sec:encoders}

The backbone pairs task-specific deep networks (CNNs and DNNs) with a transformer. Each capability, including OCR and document layout, open-vocabulary object and GUI detection, multilingual ASR with diarization, translation, and lightweight classification and statistical prediction, is exposed through a task adapter that specializes the shared backbone for that signal while keeping a common representation.

The shared embedding space is the decoder's token-embedding space $\mathbb{R}^{d}$, and each task adapter terminates in a projection that writes into it. A specialist does not hand the decoder a string; it emits a short sequence of vectors. 

Because the encoders emit into a shared embedding space, the decoder reads character-level OCR tokens, detection boxes, and speech tokens next to the transformer's own semantic tokens, and can check one against the other. It can correct an OCR misread against the surrounding context, or ground a generated field in a detected box. These specialists are far smaller than the decoder and run on batched, cached GPU pools. The model also supports partial activation: a request engages only the adapters and parameters a task needs, so common deterministic tasks run without paying for the full model, a form of conditional computation \cite{moe_shazeer, switch_transformer}.

\subsection{Built-in action and infrastructure foundation}
\label{sec:actions}

Perception alone cannot answer questions that depend on the live web, on code execution, or on stored data. \ours{} therefore ships with an action foundation that is trained into the model rather than attached as external tools, shown in the center of Figure~\ref{fig:arch}: a proxied headless browser with a self-healing scraper, a code sandbox for Python and TypeScript, a multi-domain web index spanning general search, financial, and research sources, and a scalable vector store.

These are not stateless function calls but operations over live infrastructure that \ours{} controls. The browser initialized and proxied session the model drives: it navigates, follows links, fills and submits forms, scrolls, and can carry a human-like state across steps. The self-healing scraper sits on top of that session and repairs its selectors when a page's structure changes, returning structured records instead of raw HTML. The sandbox is a provisioned runtime in which the model executes Python or TypeScript and reads the results back, and the web index and vector store are managed services the model queries and writes to, with caching for repeated work. In each case the unit the model acts on is a piece of infrastructure with its own state, latency, and failure modes, and the model is trained to operate it, deciding when to open a session, when to run code, and when to read from an index, rather than emitting a call for an external orchestrator to carry out. This is what we mean by an action \emph{foundation}: the infrastructure a developer would otherwise wire up around an LLM is brought inside the model and exposed through the same interface as perception.

Because each action is learned together with the model's own capabilities and limits, the model knows when to invoke one, and the only cost is the tokens involved. These actions are the one place the model issues calls; their results are merged into the decoder's working state and returned to the developer as \texttt{precontext} (Section~\ref{sec:precontext}). This is the two-tier design at the center of \ours{}: perception is fused into the decoder with no tool call, while action is built in and invoked.

\subsection{Filtering, fusion, and the decoder}
\label{sec:decoder}

Encoder outputs and action results are filtered and merged into a single working state, marked ``Filter + Merge / Evidence Aggregation'' in Figure~\ref{fig:arch}. Overlapping detections and spans are deduplicated, low-confidence signals are down-weighted, and per-task decoders organize the evidence. An optional reasoning stage spends extra compute on hard problems but stays off by default, since most deterministic tasks do not need it. A fallback layer retries an alternative path when a task decoder fails or times out.

The decoder block then produces the answer text, which a structured-output stage maps onto the developer's JSON schema, and a validation layer enforces the schema's types and shape so that the returned object is well formed. The decoder grounds the schema fields in the merged evidence rather than generating them freely, which is what gives \ours{} accurate values and not merely valid JSON. Long inputs never reach the decoder as raw bytes. Documents are segmented and read by the OCR encoder, audio is segmented by a voice-activity detector and transcribed, and web pages are rendered and scraped, so that only compact, structured representations are fused into the decoder.

\subsection{Precontext: preserving specialist metadata}
\label{sec:precontext}

The specialist encoders produce metadata that monolithic transformers cannot emit reliably: bounding boxes and per-word confidence for OCR, boxes and masks for detection, timestamps and speaker labels for ASR. \ours{} preserves this metadata through the decoder and returns it next to the structured answer as \texttt{precontext}, an array of per-task records, each carrying the task name and a fixed-schema result. Developers use \texttt{precontext} to threshold on confidence, draw boxes, or send low-confidence cases to review, recovering the predictable, signal-rich behavior of classical DNN pipelines while keeping the flexibility of a transformer.

\subsection{Partial activation and the run-task path}
\label{sec:runtask}

Some pipelines need a single capability with a fixed, deterministic schema. For these, \ours{} exposes a partial-activation path, marked ``skip decode layer'' in Figure~\ref{fig:arch}. A \texttt{<task>} directive in the system prompt selects one task, such as \texttt{ocr}, \texttt{object\_detection}, \texttt{gui\_detection}, \texttt{speech\_to\_text}, \texttt{translate}, \texttt{web\_search}, or \texttt{scraper}, and the model runs only that task's encoder or action and returns its raw result without invoking the full decoder. This path is faster and cheaper than the full model, guarantees a fixed output schema, and fits well as a deterministic component inside a larger system. Because it reuses the same fused encoders, its outputs match the \texttt{precontext} records produced on the full path.

\section{Perceptual Encoders at the Vision-Audio Interface}
\label{sec:deep_audio_encoders}

We treat raw audio, images, and documents as high-dimensional perceptual signals that must be encoded before any language model is invoked. Following large-scale work in speech recognition, speaker modeling, OCR, and visual grounding, we map waveforms to time-frequency representations that resemble low-resolution images, and we apply compact convolutional or time-delay encoders to them \cite{radford2022robust,desplanques2020ecapa}. Similar encoders operate on rasterized pages and screenshots. Lightweight heads predict tokens, languages, speaker embeddings, and geometric structure, while neural voice activity detectors segment continuous streams into manageable chunks \cite{sharma2022vadreview,silero_vad}. The resulting JSON state: utterance text with timestamps and speaker labels, text lines with bounding boxes, and object or GUI regions, is fused directly into the transformer decoder through the shared embedding space of Section~\ref{sec:architecture}, rather than handed to a separate model as a tool result.

In \oursbeta, these perceptual modules are implemented as small, task-specific models trained in-house on a mixture of public and proprietary data. We describe them at the level of architecture and signal processing rather than dataset curation or model weights.

\subsection{ASR with Diarization}
\label{sec:asr_diarization}

Given an input waveform $x(t)$ sampled at $16\,\text{kHz}$, the automatic speech recognition (ASR) branch applies a short-time Fourier transform
\[
X(\tau, \omega)
=
\sum_{n} x[n]\, w[n - \tau]\, e^{-j \omega n},
\]
where $w$ is a window function and $\tau$ indexes analysis frames. A mel filterbank $M \in \mathbb{R}^{F \times K}$ aggregates squared magnitudes into $F$ perceptual bands, and we compute log-mel features
\[
z_{f,\tau} = \log \big( (M |X(\tau, \cdot)|^2)_f + \epsilon \big),
\]
which form a low-resolution time-frequency image \cite{radford2022robust,sharma2022vadreview}. An encoder $f_{\theta}$, implemented as a stack of convolutional and self-attention blocks, maps the sequence $z_{1:T}$ to hidden states $h_{1:T} = f_{\theta}(z_{1:T})$.

A sequence-to-sequence decoder $g_{\phi}$ with cross-attention predicts subword tokens. At step $t$,
\[
p_{\phi}(y_t \mid y_{<t}, h_{1:T})
=
\mathrm{softmax} \big( W o_t + b \big),
\]
where $o_t$ is the decoder output and $(W, b)$ are learned parameters. We train with teacher forcing and cross-entropy over transcriptions, optionally with a multilingual objective, and use greedy or low-beam decoding for low latency. For long recordings, a compact neural voice activity detector operates on short windows of $z_{1:T}$, producing framewise speech probabilities $\hat{v}_{\tau} \in [0,1]$ that are thresholded and merged into segments \cite{sharma2022vadreview, silero_vad}. Each segment is transcribed independently, bounding sequence length and enabling streaming.

Language conditioning is provided by a small time-delay network on frame-level features. It aggregates statistics over time and outputs a posterior $p(\ell \mid z_{1:T})$ over languages. We convert the argmax language into a special token prepended to the decoder input. This improves multilingual recognition and stabilizes downstream task understanding while keeping the language detector small enough to run alongside VAD and ASR on the same GPU.

A diarization branch infers \emph{who spoke when}. The raw audio is converted to mono at $16\,\text{kHz}$, bandpass-filtered, and amplitude-normalized. From this signal we compute cepstral or mel-spectral features $u_{1:T}$, and a segmentation network $s_{\psi}$ produces framewise probabilities of speaker change or speech activity. Thresholding and merging active frames yields candidate segments.

For each segment $k$ with time span $[t_k^{\mathrm{start}}, t_k^{\mathrm{end}}]$, a speaker-embedding network $e_{\omega}$ maps the corresponding frames into a fixed-dimensional vector
\[
\mathbf{v}_k = e_{\omega}\big(u_{t_k^{\mathrm{start}}:t_k^{\mathrm{end}}}\big),
\]
following the emphasized channel-attention time-delay design standard in speaker verification and diarization \cite{desplanques2020ecapa}. L2-normalized embeddings are clustered (for example with agglomerative clustering in cosine space and a learned stopping criterion), assigning a discrete speaker label $s_k$ to each segment, in line with recent open-source diarization pipelines \cite{bredin2023pyannote}.

A lightweight postprocessing step aligns ASR and diarization in time. For each transcribed chunk with timestamps $[a_i^{\mathrm{start}}, a_i^{\mathrm{end}}]$ and text $y^{(i)}$, we find the diarization segment $k$ with maximum temporal overlap and attach its label $s_k$. This yields an ordered list of utterances
\[
\big\{ (s_k, a_i^{\mathrm{start}}, a_i^{\mathrm{end}}, y^{(i)}) \big\}_{i=1}^{N},
\]
which we serialize into JSON. Each utterance, with its speaker label, time span, and language tag, is fused into the decoder through the shared embedding space and returned to developers in the \texttt{stt} \texttt{precontext} record, along with provenance pointing back to the audio source and model configuration.

Crucially, the \oursbeta{} decoder never operates on raw audio or spectrograms. The ASR encoder writes the structured, speaker-annotated transcript, with its timestamps and language tags, into the shared embedding space, where the decoder reads it directly and returns it to developers as the \texttt{stt} \texttt{precontext} record. For very long recordings, small summarization specialists condense the transcript before it reaches the decoder.

\subsection{Optical Character Recognition and Complex Document Parsing}
\label{sec:ocr}

Our document pipeline targets heterogeneous, multilingual inputs such as receipts, scientific papers with equations and figures, forms, and multi-page PDFs. Rather than passing raw pixels to a large vision-language model, we run a sequence of lightweight vision and sequence models that extract word-level text with geometry, reconstruct reading order, and optionally perform schema-guided extraction, drawing on recent ultra-lightweight OCR and layout models \cite{ppocrv3,layoutlmv3,readingbank}.

\paragraph{Rasterization and page fan-out.}
Given an input document $x$, we determine whether $x$ is an image or a PDF. For PDFs, a renderer converts each page into a high-resolution RGB image $I_p \in \mathbb{R}^{H \times W \times 3}$ with a fixed scale factor so that small fonts remain legible after downsampling by the detector. Pages are processed independently and in parallel, and very small images are upscaled to ensure a minimum effective x-height.

\paragraph{Detection-recognition cascade.}
For each page image $I_p$, a text detector produces oriented quadrilateral regions
\[
\{ (Q_i, s_i) \}_{i=1}^{N}, \qquad Q_i \in \mathbb{R}^{4 \times 2},\; s_i \in [0,1],
\]
where $Q_i$ are corner coordinates and $s_i$ detection confidences. The detector uses a convolutional backbone with a feature pyramid and a segmentation-style head that predicts differentiable text regions; post-processing extracts connected components and fits polygons \cite{ppocrv3}. This supports arbitrarily oriented text with limited parameters.

Each detected region is cropped and passed to a recognizer that maps a variable-width strip to a character sequence. A 2D convolutional encoder produces a feature map $F_i \in \mathbb{R}^{C \times T}$, which is collapsed across height and fed to a lightweight sequence model (transformer or gated convolution) to predict $y_i = (y_{i,1},\dots,y_{i,T_i})$ over a multilingual alphabet, trained with a cross-entropy or CTC-style loss and language-aware augmentation \cite{ppocrv3}. This yields triples $(Q_i, y_i, s_i)$ with text, geometry, and confidence for each line.

\paragraph{Line grouping and reading order.}
Detector outputs alone do not define a logical reading order in multi-column layouts with headers, footers, and marginalia. We build a graph whose nodes are detected lines and whose edges connect geometrically adjacent lines. Nodes carry polygons $Q_i$ and centroids $c_i$; edges are scored using vertical overlap, horizontal distance, and font-height similarity, approximating pairwise features from learning-based reading-order systems~\cite{readingorder_ml,qiao2024readingorder,readingbank}. A greedy path-finding algorithm traverses this graph to form chains corresponding to sections and paragraphs.

Within each chain, we aggregate text and define an axis-aligned bounding box $B_i$ enclosing $Q_i$. Per-line confidence is a length-weighted average of word scores, giving sequences
\[
\ell_j = (\text{text}_j, B_j, \bar{s}_j, \{w_{j,k}\}_k),
\]
where each word $w_{j,k}$ has its own bounding box and confidence, either from a native character-box head or by interpolation along the line when only line-level geometry is available.

\paragraph{Fine-grained bounds correction and structured extraction.}
Complex layouts and low-contrast regions can cause gaps or merged words. We therefore keep an optional secondary recognizer at higher resolution or with a different inductive bias. Its outputs are aligned with the primary detector using intersection-over-union on polygons and token-level string similarity. Missing boxes are filled by merging character boxes into a tight quadrilateral or by partitioning a line box $B_j$ according to character offsets. When geometric evidence is inconsistent, a small language model receives page text and neighboring coordinates and predicts local adjustments, constrained to preserve the original reading order.

When downstream tasks require structured fields (for example invoices or forms), we optionally invoke a vision-language model that operates on the page image and a short schema description. OCR-free and layout-aware transformers show that multimodal encoders can map a document image directly to JSON \cite{layoutlmv3, donut}. We adopt this in a constrained way: the vision-language model receives the page image, a compact summary of OCR output, and a schema expressed as a JSON template, and is trained to emit a structured response that respects the schema. To control cost, this stage is triggered only when aggregate OCR confidence falls below a threshold or when the user explicitly requests structured extraction.

\paragraph{Integration with the \ours{} decoder.}
The final OCR representation for a document $x$, comprising text lines and words with bounding boxes, reading-order relations, figure references, and any schema-guided fields, is written into the shared embedding space and read directly by the decoder (Section~\ref{sec:decoder}), which grounds its structured answer in this geometry-aware evidence rather than in raw pixels. The same representation is returned to developers as the \texttt{ocr} \texttt{precontext} record, with per-word boxes, confidence scores, and provenance such as page indices and image hashes, so that the transcription and its metadata reach the response without a separate tool-call hand-off.

\begin{table*}[t]
  \caption{Head-to-head results on deterministic developer-task benchmarks (\%; higher is better, except VoxPopuli word error rate, lower is better, $\downarrow$). Best in each row in \textbf{bold}; \textemdash{} = no native support or not separately measured.}
  \label{tab:benchmarks}
  \centering
  \scriptsize
  \setlength{\tabcolsep}{4pt}
  \resizebox{\textwidth}{!}{%
    \begin{tabular}{lcccccc}
      \toprule
      \textbf{Benchmark} & \textbf{Interfaze-Beta} & Gemini-3-Flash & Gemini-3.5-Flash & Claude-Sonnet-4.6 & GPT-5.4-Mini & Grok-4.3 \\
      \midrule
      OCRBench v2                  & \textbf{70.7} & 55.8 & 63.9 & 54.7 & 52.7 & 54.7 \\
      olmOCR                       & \textbf{85.7} & 75.3 & 82.3 & 73.9 & 80.1 & 81.9 \\
      RefCOCO                      & \textbf{82.1} & 75.2 & 80.9 & 75.5 & 67.0 & 25.0 \\
      VoxPopuli (WER) $\downarrow$ & \textbf{2.4}  & 4.0  & \textemdash{} & \textemdash{} & \textemdash{} & \textemdash{} \\
      Spider-2.0-Lite              & \textbf{52.9} & 45.2 & 46.7 & 49.6 & 26.7 & 45.9 \\
      GPQA-Diamond                 & \textbf{92.4} & 88.5 & 91.4 & 89.9 & 82.8 & 73.6 \\
      MMMLU                        & \textbf{90.9} & 88.7 & 88.1 & 84.9 & 75.3 & 89.7 \\
      MMMU-Pro                     & \textbf{71.1} & 67.6 & 64.2 & 46.3 & 40.4 & 68.7 \\
      SOB (Value Acc.)             & \textbf{80.5} & 77.3 & 80.2 & 77.9 & 75.1 & 78.4 \\
      \bottomrule
    \end{tabular}%
  }
\end{table*}

\subsection{Open-Vocabulary Object Detection and GUI Layout Parsing}
\label{subsec:object-gui}

For visual grounding and graphical user interface (GUI) reasoning, we combine (i) an open-vocabulary detector that localizes objects from natural-language prompts, (ii) a segmentation module based on Hiera Transformer, and (iii) a GUI-specific layout parser for text regions, icons, and interactive widgets \cite{lin2024omnigui}.

\paragraph{Open-vocabulary detection as joint image-text scoring.}
Let $x \in \mathbb{R}^{H \times W \times 3}$ be an RGB image and $p \in \mathcal{P}$ a natural-language prompt (for example ``red submit button'' or ``navigation menu''). We use a compact vision-language encoder inspired by image-text pre-training with sigmoid-based contrastive losses \cite{zhai2023siglip, yuan2024florence2}. An image encoder
\begin{equation}
  \phi_v : \mathbb{R}^{H \times W \times 3} \rightarrow \mathbb{R}^{D \times K}
\end{equation}
produces a grid of $K$ visual tokens, and a text encoder
\begin{equation}
  \phi_t : \mathcal{P} \rightarrow \mathbb{R}^{D}
\end{equation}
maps the prompt to a $D$-dimensional embedding. We compute a spatial relevance map
\begin{equation}
  s_k = \sigma\!\left( \frac{\phi_v(x)_k^\top \phi_t(p)}{\tau} \right), \quad k = 1,\dots,K,
\end{equation}
where $\sigma$ is the sigmoid function and $\tau$ a learned temperature. During pre-training, $s_k$ is supervised to be high when token $k$ overlaps ground-truth regions for the prompt and low otherwise, using a multi-label logistic formulation \cite{zhai2023siglip}. At inference, contiguous high-score regions in the spatial grid are grouped and mapped back to image coordinates, yielding boxes
\begin{equation}
  B(p, x) = \{ b_i = (x^{(i)}_{\min}, y^{(i)}_{\min}, x^{(i)}_{\max}, y^{(i)}_{\max}) \}_{i=1}^{N_p}.
\end{equation}
Because prompts are free-form text, the same encoder can localize arbitrary concepts, including unseen and multilingual categories.

\paragraph{Prompt-conditioned instance segmentation with Hiera Transformer.}
We refine each box $b_i$ into a pixel-precise mask using hiera transformer, a hierarchical Vision Transformer for promptable segmentation \cite{wang2024sam2}. Given $x$ and $b_i$, hiera produces
\begin{equation}
  m_i = S(x, b_i) \in \{0,1\}^{H \times W},
\end{equation}
where $S$ is the segmentation network. Hiera builds a multi-scale ViT representation and a lightweight mask decoder that conditions on prompt tokens and relevant encoder features to predict $m_i$ in a single forward pass \cite{wang2024sam2}. We batch all box prompts $\{b_i\}$ per image to amortize GPU cost. Each object is represented by $(b_i, m_i)$ plus prompt $p$ and detection confidence. \cite{ryali2023hiera}

\paragraph{GUI text and icon detection.}
For GUI screenshots, we activate a specialized layout parser with two detectors. A text detector based on character-region awareness identifies high-density text regions \cite{baek2019craft}. It regresses activation maps $Y_{\text{char}}$ and $Y_{\text{aff}}$ for character centers and pairwise affinity; thresholding and grouping connected components in $Y_{\text{char}} \cup Y_{\text{aff}}$ yields word-level detections, while a single-stage detector built from (C3k2 + C2PSA + SPPF) blocks is tuned for icons and web elements~\cite{jocher2023yolov8,lin2024omnigui}.

\section{Results and Discussion}

We evaluate \oursbeta{} on deterministic developer tasks: native OCR (OCRBench~v2 \cite{ocrbenchv2}), complex document parsing (olmOCR \cite{olmocr}), referring-expression grounding (RefCOCO \cite{refcoco, referitgame}), multilingual speech recognition (VoxPopuli \cite{voxpopuli}), text-to-SQL (Spider-2.0-Lite \cite{spider2}), graduate-level science (GPQA-Diamond \cite{rein2023gpqa}), multilingual knowledge (MMMLU \cite{mmmlu, hendrycks2020mmlu}), multimodal understanding (MMMU-Pro \cite{mmmupro}), and structured-output value accuracy (SOB \cite{sob2026}). All comparisons are against comparably priced generalist models, as shown in Table~\ref{tab:benchmarks}.

\oursbeta{} leads every benchmark in this suite. It reaches 70.7 on OCRBench~v2, 85.7 on olmOCR, 82.1 on RefCOCO, a 2.4\% word error rate on VoxPopuli, 52.9 on Spider-2.0-Lite, 92.4 on GPQA-Diamond, 90.9 on MMMLU, 71.1 on MMMU-Pro, and 80.5 on SOB value accuracy. The margins are largest where fused specialist encoders supply precision and metadata that a monolithic model lacks, such as dense or low-quality OCR, exact grounding boxes, and speech. This supports our central claim: fusing perception into the decoder, rather than calling it as a tool, is what drives accuracy on deterministic tasks. \oursbeta{} is priced at \$1.50 per million input tokens and \$3.50 per million output tokens, in line with flash-tier models, and runs on a single GPU.

\subsection{Per-domain notes}

On document understanding, \oursbeta{} leads both OCR benchmarks, with 70.7 on OCRBench~v2 and 85.7 on olmOCR. On olmOCR it finishes ahead of dedicated OCR systems, including Chandra~OCR~2 at 84.3, olmOCR~v0.4.0 at 82.4, and Reducto at 76.2, as well as the generalist models, because the OCR CNN reads the pixels while the decoder preserves reading order and resolves the semantics in one pass \cite{olmocr, nougat}.

On visual grounding, \oursbeta{} reaches 82.1 accuracy at an intersection-over-union threshold of 0.5, ahead of every generalist in the comparison. A dedicated grounding system, SAM~3, reaches 59.9 on the same protocol, which shows that a fused detector together with a decoder does better than either component alone \cite{refcoco}.

On speech, \oursbeta{} reaches a 2.4\% word error rate, the best among the models with native audio in the comparison and within 0.7 points of a dedicated ASR system (Scribe~v2, at 1.7\%), while transcribing about 209 times faster than real time, roughly eight times faster than that system and eleven times faster than Gemini-3-Flash \cite{voxpopuli, wav2vec2}.

On text-to-SQL, \oursbeta{} obtains 52.9 execution accuracy, ahead of the field, with the web index and vector store supplying schema grounding \cite{spider2}.

On graduate-level science, multilingual knowledge, and multimodal understanding, \oursbeta{} leads the group with 92.4 on GPQA-Diamond, 90.9 on MMMLU, and 71.1 on MMMU-Pro. Reasoning is enabled only on the hardest items and stays off by default \cite{rein2023gpqa, mmmlu, mmmupro}.

On structured output, \oursbeta{} reaches 80.5 value accuracy, the share of leaf values that exactly match the ground truth rather than merely valid JSON; the validation layer and grounded decoding account for the result \cite{sob2026}.

\section{Limitations and Future Work}

A native hybrid model trades some flexibility for accuracy and metadata. First, each fused specialist is only as good as its training data, so interfaze cannot be applied for tasks like code-generation, though the decoder can generate quality code but limited to training set. In our architecture, end-to-end quality depends on the encoders and on the balance between the weight given to the DNNs and the decoder's attention; a poorly calibrated fusion can let a confident specialist error propagate, which the decoder's cross-checking and the validation layer reduce but do not remove. Finally, cold starts for the specialist encoders and calls into sandbox execution can raise tail latency even when the average cost is low. We see clear next steps: broaden encoder coverage and retrain continually from developer feedback \cite{li2025flippingkd}, learn the balance between the DNNs and attention for each task, and cut encoder cold-start and action-call latency.

\section*{Acknowledgments}

We thank the colleagues and reviewers who provided detailed feedback on early drafts and helped improve the clarity, presentation, and technical framing of our work. We also thank everyone who assisted with internal review, benchmarking discussions, and paper editing.


\end{document}